\documentclass{article} 
\usepackage{paper,times}

\usepackage{amsmath,amsfonts,bm}

\def\eqref#1{equation~\ref{#1}}

\def\1{\bm{1}}

\DeclareMathAlphabet{\mathsfit}{\encodingdefault}{\sfdefault}{m}{sl}
\SetMathAlphabet{\mathsfit}{bold}{\encodingdefault}{\sfdefault}{bx}{n}

\usepackage{hyperref}
\usepackage{url}
\usepackage{booktabs}
\usepackage{amsmath}
\usepackage{graphicx}
\usepackage{multirow}
\usepackage{amssymb}
\usepackage{amsthm}
\newtheorem{proposition}{Proposition}

\newtheorem{lemma}{Lemma}

\usepackage{pgfplots}
\pgfplotsset{compat=1.18} 
\usepgfplotslibrary{groupplots}
\usepackage{wrapfig}
\usepackage{float}

\title{Learn the Solid, Not the File: Canonical Inputs for Neural Networks on CAD Boundary Representations}

\iclrfinalcopy

\author{Heinrich Jiang,  Hager Yasser Mohamed, Alexander Hitt, Valeriia Lomakina,\\
\textbf{Henning Jiang, Jennifer Jang} \\
StoryGold AI \\
\texttt{\{heinrich,hager,alex,valeriia,henning,jennifer\}@storygold.com}
}

\begin{document}

\maketitle

\begin{abstract}
Boundary representation (B-rep) is the standard format used by modern CAD systems for parametric 3D models. It turns out, the exact same solid can be represented by different B-reps: for example, two engineers using different operations, a geometry kernel rebuilding the file, and an export setting repartitioning faces will lead to different B-reps even though the underlying solid remains the same.

We show that existing B-rep encoders are not robust to variation in the B-rep with the same solid on perturbations applied to standard benchmarks, naturally occurring variations inherent to CAD software, and differences in how designers model the same part via a human dataset we created in FreeCAD. The performance of popular B-rep encoders often collapses catastrophically.

We propose the \emph{canonical region graph}, an input representation whose nodes, features and coordinate frame are derived from the solid itself and show theoretical invariance guarantees on repartitioning and rigid motions.  It matches the strongest baseline on standard benchmarks, and is stable under every perturbation we test.
\end{abstract}

\section{Introduction}

Computer Aided Design (CAD) is a long-standing technology that has had a profound influence on how just about any object was designed, from everyday consumer goods to architectural infrastructure and aerospace vehicles \citep{groover1983cad}. Boundary representations (B-reps) are the most popular and standard format for most modern CAD software \citep{weiler1986topological}. A B-rep describes a solid by its bounding surface patches (faces), the curves where patches meet (edges), and their
connectivity. It gives the precise mathematical instructions to create the object \citep{stroud2006boundary}.

A growing body of work trains neural networks directly on B-reps and have been used for wide range of applications such as machine feature recognition \citep{yao2026brepmae}, semantic segmentation \citep{lou2023brep}, part classification \citep{li2026masked}, CAD retrieval \citep{usama2026brepclip}, generative design \citep{jayaraman2022solidgen}, CAD synthesis \citep{xu2024brepgen}, CAD sequence reconstruction \citep{zhang2024brep2seq}, manufacturability and cost estimation \citep{ballegeer2026cad}, assembly and joint prediction \citep{willis2022joinable} and engineering simulation \cite{heidari2025geometric}.

It turns out that the B-rep is only one of many valid ways to represent the same solid. There are many different decompositions of the same surface: the same geometric boundary can admit arbitrarily many valid topological decompositions into faces and edges \citep{tierney2017using}, placed in any coordinate frame with rigid motions, and its surfaces written as analytic primitives or as splines (ISO 10303-42, \citeyear{iso10303-42}). An export setting in the CAD software itself can re-partition a solid's boundary. Moreover, the final B-rep often is decided by factors no mechanical engineer thinks are conscious design decisions (e.g. the order of modeling operations, whether a profile was sketched as one curve or two, a cleanup checkbox, and which geometry kernel produced or last translated the file) \citep{park2021boundary,zhang2020approach}. Two engineers modeling the same part, or one file passing through two CAD systems, routinely yield different drawings of an identical solid. As such, every solid admits an \emph{equivalence class} of B-reps. 

Therefore, one should naturally expect that a B-rep neural network, should be {\it robust} to the equivalence class of B-reps. That is, any valid B-rep of the solid should produce the same output, or at least approximately so. To our knowledge, none of the existing preprocessing methods enforce such robustness; and moreover, we show on standard benchmark datasets that these models often significantly break down in performance when such valid perturbations of the B-rep are applied.
Despite the widespread use and the amount of research interest in these B-rep neural networks, such a lack of robustness surprisingly remains.

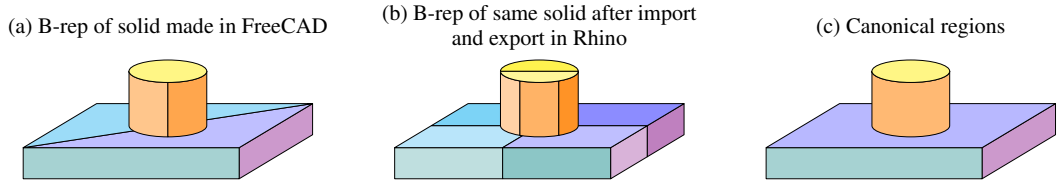
\begin{figure}[t]
\centering
\resizebox{\linewidth}{!}{%
\begin{tikzpicture}[scale=0.62, line join=round, line cap=round, font=\small]
\begin{scope}
  \fill[violet!40,draw=black,thin] (5,0)--(6.7,1.02)--(6.7,1.74)--(5,.72)--cycle;
  \fill[teal!35,draw=black,thin] (0,0)--(5,0)--(5,.72)--(0,.72)--cycle;
  \fill[blue!25,draw=black,thin] (0,.72)--(5,.72)--(6.7,1.74)--cycle;
  \fill[cyan!35,draw=black,thin] (0,.72)--(6.7,1.74)--(1.7,1.74)--cycle;
  \fill[orange!45,draw=black,thin] (2.45,2.49) arc[start angle=180,end angle=270,x radius=.9,y radius=.26]
      -- (3.35,.97) arc[start angle=270,end angle=180,x radius=.9,y radius=.26] -- cycle;
  \fill[orange!70,draw=black,thin] (3.35,2.23) arc[start angle=270,end angle=360,x radius=.9,y radius=.26]
      -- (4.25,1.23) arc[start angle=360,end angle=270,x radius=.9,y radius=.26] -- cycle;
  \fill[yellow!60,draw=black,thin] (3.35,2.49) ellipse [x radius=.9, y radius=.26];
  \node[align=center] at (3.35,3.5) {(a) B-rep of solid made in FreeCAD};
\end{scope}
\begin{scope}[shift={(8.6,0)}]
  \fill[violet!30,draw=black,thin] (5,0)--(5.85,.51)--(5.85,1.23)--(5,.72)--cycle;
  \fill[violet!50,draw=black,thin] (5.85,.51)--(6.7,1.02)--(6.7,1.74)--(5.85,1.23)--cycle;
  \fill[teal!25,draw=black,thin] (0,0)--(2.5,0)--(2.5,.72)--(0,.72)--cycle;
  \fill[teal!45,draw=black,thin] (2.5,0)--(5,0)--(5,.72)--(2.5,.72)--cycle;
  \fill[cyan!25,draw=black,thin] (0,.72)--(2.5,.72)--(3.35,1.23)--(.85,1.23)--cycle;
  \fill[blue!25,draw=black,thin] (2.5,.72)--(5,.72)--(5.85,1.23)--(3.35,1.23)--cycle;
  \fill[cyan!45,draw=black,thin] (.85,1.23)--(3.35,1.23)--(4.2,1.74)--(1.7,1.74)--cycle;
  \fill[blue!45,draw=black,thin] (3.35,1.23)--(5.85,1.23)--(6.7,1.74)--(4.2,1.74)--cycle;
  \fill[orange!35,draw=black,thin] (2.45,2.49) arc[start angle=180,end angle=240,x radius=.9,y radius=.26]
      -- ++(0,-1.26) arc[start angle=240,end angle=180,x radius=.9,y radius=.26] -- cycle;
  \fill[orange!60,draw=black,thin] ({3.35+.9*cos(240)},{2.49+.26*sin(240)}) arc[start angle=240,end angle=300,x radius=.9,y radius=.26]
      -- ++(0,-1.26) arc[start angle=300,end angle=240,x radius=.9,y radius=.26] -- cycle;
  \fill[orange!85,draw=black,thin] ({3.35+.9*cos(300)},{2.49+.26*sin(300)}) arc[start angle=300,end angle=360,x radius=.9,y radius=.26]
      -- (4.25,1.23) arc[start angle=360,end angle=300,x radius=.9,y radius=.26] -- cycle;
  \fill[yellow!50,draw=black,thin] (2.45,2.49) arc[start angle=180,end angle=360,x radius=.9,y radius=.26] -- cycle;
  \fill[yellow!80,draw=black,thin] (4.25,2.49) arc[start angle=0,end angle=180,x radius=.9,y radius=.26] -- cycle;
  \node[align=center] at (3.35,3.5) {(b) B-rep of same solid after import\\ and export in Rhino};
\end{scope}
\begin{scope}[shift={(17.2,0)}]
  \fill[violet!40,draw=black,thin] (5,0)--(6.7,1.02)--(6.7,1.74)--(5,.72)--cycle;
  \fill[teal!35,draw=black,thin] (0,0)--(5,0)--(5,.72)--(0,.72)--cycle;
  \fill[blue!28,draw=black,thin] (0,.72)--(5,.72)--(6.7,1.74)--(1.7,1.74)--cycle;
  \fill[orange!55,draw=black,thin] (2.45,2.49) arc[start angle=180,end angle=360,x radius=.9,y radius=.26]
      -- (4.25,1.23) arc[start angle=360,end angle=180,x radius=.9,y radius=.26] -- cycle;
  \fill[yellow!60,draw=black,thin] (3.35,2.49) ellipse [x radius=.9, y radius=.26];
  \node[align=center] at (3.35,3.5) {(c) Canonical regions};
\end{scope}
\end{tikzpicture}}
\caption{Canonical Regions. We show the face partitionings of the same solid that can naturally arise and how the canonical regions re-partitions the solid. (a) A B-rep that was made in FreeCAD. (b) The same solid after importing it and exporting it in Rhino 3D CAD software. 
 (c) Canonical regions of this solid which is used to construct the canonical region graph. Both (a) and (b) therefore would lead to the same graph using our method, while for the baselines, the graph depends on the original partitioning.}
\label{fig:concept}
\end{figure}


We present a new graph-based B-rep input representation called {\it canonical region graph}. There are two key components. The first is {\it canonical regions}: undo the arbitrary cuts: wherever several faces are pieces of the same underlying surface (e.g. multiple arcs of one cylinder, two halves of one plane) and merge them, so that a node of our graph is a whole surface region, not whichever fragment of it the file happened to contain; two nodes are connected when their corresponding regions
touch. The second is {\it canonical frames}: orient coordinate system in a frame from the solid itself, that is, the origin at its center of mass, and axes along its principal directions. Thus, the frame is invariant to rotations, rigid motions, and coordinate system of the file. Then, the features computed along these canonical regions and boundaries become provably invariant to repartitioning and rigid motions.

We provide a study showing that popular B-rep encoders are not invariant to this equivalence class of B-rep models for the same solid. We demonstrate this on two kinds of perturbations. The first is
automatic: to every benchmark test solid, we apply re-partitioning of faces by axis-aligned planes, random rigid motion, re-expression of every analytic surface as its exact NURBS equivalent the
operation kernel translators perform, and a 
plain import/export round-trip through a commercial NURBS-native kernel on
the entire test split. The second is human: a dataset we created and labeled by CAD experts in FreeCAD. The dataset was labeled independently by two CAD experts, with one of the experts giving three different ways of modeling the same solid so that we capture both variations possible from one labeler and variations across labels. In all cases, the popular B-rep encoders collapse in performance when such perturbations are introduced, while our method is not only robust but matches or exceeds the compared methods in performance on the standard benchmarks.

\section{Related Work}

\paragraph{Learning on B-reps.} 

There has been a lot of interest in encoder models on B-reps \citep{liu2025hola,dai2025brepformer,liang2025cadcl,kim2026brepcoder,jiang2026masked}. 
A native B-rep cannot be used directly as input for a standard neural network: it must be first preprocessed into a tractable input format. The field has consolidated to a handful of approaches to turn a B-rep into a neural network input and each inherits specific properties of the B-rep that may not be relevant the underlying solid itself.
UV-Net \citep{jayaraman2021uv} attaches to every face a small
image grid sampled in the face's own UV parameterization (and to every
edge a curve grid), encodes both convolutionally, and passes messages over
the face-adjacency graph: the features therefore inherit the
parameterization of the faces. BRepNet \citep{lambourne2021brepnet} defines convolution directly on the coedge structure with learned kernels expressed as topological walks. The topology is from the B-rep, so
splitting or merging a face rewrites the very walks the kernels are
defined over. AAGNet \citep{wu2024aagnet} combines per-face UV grids with face and edge attributes for semantic and instance machining-feature
recognition. All three express geometry in the file's coordinate frame, so a rigid motion moves every feature. Other approaches include tokenizing the hierarchical tree of the B-rep and using the sequence as the input \citep{zhang2024brep2seq}, which directly depends on the hierarchical tree of the B-rep, and reducing the B-rep down to a list of primitives \citep{wu2021deepcad}; however, this can only be used on very primitive CADs. Our canonical graph region method is a new input that focuses on the information about the solid and avoids using information that's particular to how the B-rep was constructed.

\paragraph{Robustness for B-rep Neural Networks.} The UV-Net work
observed a piece of the problem \citep{jayaraman2021uv}, that a
face's UV grid can be sampled in several equivalent orders and plain grid
convolutions are sensitive to the choice, for which they propose
to fix the sampling of a \emph{fixed} face on a fixed partition of the solid's boundary into faces; but it does not address changes to the partitioning, and both the UV-Net and AAGNet inherit this. BRepGAT \citep{lee2023brepgat} mentions a large gap in performance on between their own dataset and on externally authored models, and attributes it to
topological structure differing across CAD systems; however, they don't provide any solutions. Some recent work improve B-rep output validity and error reduction \citep{hafez2023robust,shen2025mesh2brep,liu2026dualbrep,qin2026brep,qi2026pointer,qin2026autoregressive}. \cite{lee2025replacing} addresses one source of B-rep representation variability by replacing NURBS surfaces with analytic primitives and \cite{jones2023b} learns entity correspondence under topology changes. \cite{sun2010approach} simplifies B-reps by removing unwanted features. \cite{ballegeer2026fov} proposes a rotation-invariant method.

\section{The Canonical Region Graph}
\label{sec:method}

\begin{figure}[t]
\centering
\resizebox{0.8\linewidth}{!}{%
\begin{tikzpicture}[scale=0.9, font=\small, >=stealth]
\fill[cyan!22,draw=black!70] (0,0) rectangle (1.2,2.2);
\fill[orange!25,draw=black!70] (1.2,0) -- (3.1,0) -- (2.6,2.2) -- (1.2,2.2) -- cycle;
\fill[green!22,draw=black!70] (3.1,0) -- (4.5,0) -- (4.5,2.2) -- (2.6,2.2) -- cycle;
\draw[black,very thick] (0,0) rectangle (4.5,2.2);
\foreach \cx/\cy/\a in {0.6/1.45/15, 2.0/0.75/-30, 3.7/1.25/60}{
  \begin{scope}[shift={(\cx,\cy)},rotate=\a]
    \draw[->,thick,red!35] (0,0)--(0.62,0);
    \draw[->,thick,blue!35] (0,0)--(0,0.44);
    \fill[black!45] (0,0) circle (0.04);
  \end{scope}}
\node[font=\scriptsize,align=center] at (2.25,-0.55)
  {three stored faces (any shapes),\\ three unrelated local frames};
\node[font=\Large] at (5.6,1.1) {$\Rightarrow$};
\begin{scope}[shift={(6.7,0)}]
  \fill[blue!10,draw=black,very thick] (0,0) rectangle (4.5,2.2);
  \fill[black] (2.25,1.1) circle (0.06);
  \draw[->,very thick,red!75!black] (2.25,1.1)--(3.55,1.1);
  \draw[->,very thick,blue!70!black] (2.25,1.1)--(2.25,2.0);
  \node[font=\scriptsize,align=center] at (2.25,-0.55)
    {one region, one frame:\\ centroid $+$ principal axes of the whole};
\end{scope}
\end{tikzpicture}}
\caption{Canonical Frame. Left: Each
stored face carries its own local frame (and parameterization), all of
which change when the file is re-cut or re-expressed. Right: we use a single canonical frame computed from the whole which cannot change, because the whole does not.}
\label{fig:frame}
\end{figure}
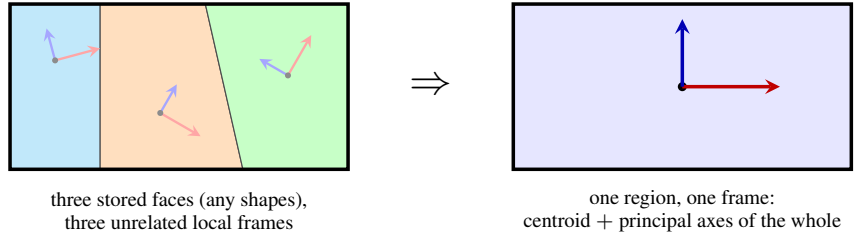

Like many B-rep encoding methods, our model consumes a graph and
is trained with a graph neural network. Existing methods typically use a graph whose nodes are the faces that are stored in the file, with features sampled from that face's parameterization in the file's coordinate frame. Our graph replaces all of this with quantities that can be computed from the solid itself: one node per \emph{canonical region} and coordinate system computed relative to the solid's own frame rather than the file's.
The full detailed construction of what we will discuss here is defined end-to-end, along with keys, tolerances, frame, grid, and every constant in Appendix~\ref{app:construction}.

{\bf Nodes and Edges}: In the canonical region graph, the nodes are the maximally connected regions after undoing the arbitrary cuts (i.e. merging faces that are pieces of one underlying surface). In order to do this, each face gets an exact key
identifying its surface in the following way: if the face is an analytic surface (a plane, cylinder, sphere, cone or torus), then the key is based on the surface type and parameters defining it. Next, if the face is saved as a spline (i.e. in NURBS, CAD's generic freeform format that defines a surface using a grid of control points rather than an equation) but actually an analytic surface up to a tolerance, the key is based on the analytic surface. The same applies when the file describes the surface procedurally rather than by shape (e.g. a straight line revolved about an axis, or a circle dragged along a direction), which is a third way a kernel can write down the identical cylinder or plane. Recovering analytic surfaces from freeform representations, and preferring the
simplest primitive that fits, is standard practice in reverse
engineering~\citep{varady1997reverse}; ours is a deterministic least-squares variant with the selection rule and its failure cases in
Appendix~\ref{app:edge}. If it's genuinely a complex NURBs surface, the key is based on the control grid, read from the underlying {\it untrimmed} surface. In a B-rep, a face in this case is stored as a bounded portion of a larger underlying surface, and splitting a face only redraws the bounds. Both pieces still point at the same underlying surface, so the key survives any cutting. It is important to note here that the key is computed based on the underlying {\it untrimmed} surface the face inherits from (as the B-rep contains this information), rather than only the bounded portion defining the face. Two faces merge exactly when their keys are equal \emph{and} they share an edge. The shared-edge condition is required because sometimes two separate features are machined to the same surface (e.g. two identical sub-parts) must still remain two separate regions, and not merge into one. These maximally connected faces become the canonical region and becomes a node in the canonical region graph. Two nodes share an edge if the corresponding canonical regions share a boundary curve.

\paragraph{The Canonical Frame.} Several features to be consumed by the graph neural network depend on positions and directions. A coordinate is only meaningful relative to a \emph{frame}. The B-rep's frame is arbitrary because it depends on how the designer oriented the part. We compute one from the solid instead: the origin is the boundary's centroid, and the axes are its principal directions. Each axis's sign is chosen by which side of the perpendicular plane through the centroid carries more boundary area (i.e. a third-moment "lopsidedness"); when a part is too symmetric for this to decide, the two choices are smoothly blended by the lopsidedness weight in order to have continuity. In cases where the principal axis directions themselves are ambiguous, we average feature values across the ambiguous rotations, in closed form. At non-degenerate configurations, the frame transforms equivariantly under rigid motions. At symmetric configurations, where no unique frame exists, we instead construct averaged features that are independent of the ambiguous basis. Keeping mirror images distinct requires some additional care in the frame construction; see Appendix~\ref{app:rigid}. Lastly, scale: before any feature is computed, the solid is uniformly rescaled so that its total boundary area is $1$. As a consequence, our method is also scale-invariant, a property the popular baselines already have \citep{jayaraman2021uv,wu2024aagnet}.

\paragraph{Node and Edge Features.} Each node carries $115$ features summarizing its region and each edge has $10$ features summarizing the shared boundary between two regions, with every position and
direction expressed in the canonical frame. The features are largely the attributes the field has always used such as surface type, area and convexity \citep{joshi1988graph} and overlap with that of BRepNet almost item for item \citep{lambourne2021brepnet}. What
differs is where they are computed: regions in the canonical frame and not the stored faces in the B-rep's frame. Thus, every feature inherits the desired invariance properties from its region and frame. In particular, there are no UV grids or features that read a parameterization.

\paragraph{Learning on the graph.} The network is a standard graph
transformer: standardized features pass through an MLP stem, $8$ rounds
of edge-conditioned multi-head attention with residual feed-forward
blocks (width $512$, $8$ heads, $17.5$M parameters), and a linear head
producing per-region logits. On tasks that require us to make a prediction based on a B-rep face, the face simply inherits its region's
prediction, so any two faces
of one region agree by construction. Exact details in
Appendix~\ref{app:construction}.

\section{Invariance guarantees}
\label{sec:theory}

We show a number of guarantees for the canonical region graph. Throughout, $S$ denotes the solid and $\partial S$ its boundary
surface. We call a partition of $\partial S$ into faces \emph{valid} if every face is contained in a single supporting parametric surface (i.e. patches do not straddle distinct surfaces) and distinct faces meet only along piecewise-regular curves and isolated vertices (i.e. finitely many smooth arcs, excluding degenerate intersections). These conditions hold for the valid B-reps: each B-rep face is a trimmed region of a single supporting surface, and face adjacencies are represented by finitely many topological edges supported by curves. Inputs violating the required B-rep validity conditions are rejected during intake (ISO 10303-42, \citeyear{iso10303-42}).

The first result shows that valid B-reps composed of the supported analytic surfaces of the same solid yield the same decomposition into canonical regions. All proofs are in Appendix~\ref{app:proofs}.

\begin{proposition}[Canonicality]\label{prop:canon}
$\partial S$ decomposes uniquely
into maximal connected patches of analytic surfaces. For \emph{every} valid
partition $P$, whose faces lie on analytic surfaces (natively or recognized at machine precision) the region graph $G(P)$ is isomorphic (with identical node
geometry) to the graph of these patches. $G$ is therefore a function of the solid, not of $P$.
\end{proposition}

The next result shows that totals computed from the pieces of the same surface equals the total over the entire surface. This makes features computable from any file and, applied to the whole boundary, makes the canonical frame itself partition-invariant.

\begin{proposition}[Additivity over subdivisions]\label{prop:integral} Surface and edge integrals are additive over subdivisions. Therefore, subdividing any region into valid subregions does not change the resulting features, so they are invariant to the choice of valid partition $P$ of $\partial S$. \end{proposition}

Next, we show that a partition-invariant model can use only the information from the shape.

\begin{proposition}[Lower bound on partition-invariance]\label{prop}
A predictor is partition-invariant if and only if its prediction depends only on
the underlying solid geometry. Consequently, for any label $Y$ (e.g., a
modeling operation), the risk of a partition-invariant predictor is bounded
below by the irreducible disagreement of $Y$ among B-reps representing the
same geometry. Thus, any predictor achieving lower risk must exploit
information in the B-rep partition that is not determined by the geometry.
\end{proposition}

Some features (e.g. the integrated normal, the curvature totals, the
dihedral histograms, the occupancy grid) cannot be computed in closed
form and they are evaluated by \emph{quadrature} on a triangle mesh of the
surface; that is the integral approximated as a finite sum, the integrand
evaluated at one sample point per triangle in the mesh and weighted by that
triangle's area. The next result guarantees that such features remain invariant up to tolerance proportional to the mesh error.

\begin{lemma}[Stability of the mesh-computed features]\label{lem:stability}
The following holds for any mesh-computed feature that is Lipschitz continuous in the quadrature sample positions and area weights. A mesh perturbation of size
$\varepsilon$ (i.e. a re-triangulation of the same surface that changes
each sample position and area weight by at most $\varepsilon$)
changes every such feature by at most $O(\varepsilon)$.
\end{lemma}

The final result shows that under the canonical frame, the features are invariant under rigid motion.

\begin{proposition}[Rigid-motion invariance]\label{prop:rigid}
The centroid and principal-axis frame transform equivariantly under rigid
motions. Consequently, features expressed in this canonical frame (after the sign blending and degenerate-subspace averaging, which are
functions of invariants alone) are exactly invariant to translation
and rotation, and remain partition-invariant.
\end{proposition}

\begin{table}[t]
\centering
\caption{Invariance under perturbations on MFInstSeg test solids. For each class of perturbations, we compare the perturbed version with the original in whether the region graph structure match and by how much the largest feature differs as a multiple of the standard deviation of that feature in the training set. We show the $50$ and $99$ percentiles and worst number among the solids.}
\label{tab:verify}
\begin{tabular}{lcccc}
\toprule
& & \multicolumn{3}{c}{exact-feature residual ($\sigma$)} \\
perturbation & identical region graph & p50 & p99 & worst \\
\midrule
diag-split (re-partition) & $99.9\%$ & $0$ & $4\!\times\!10^{-7}$ & $0.002$ \\
NURBS re-expression       & $100\%$  & $0.0007$ & $0.009$ & $0.022$ \\
random rigid motion             & $100\%$  & $0$ & $0.002$ & $0.007$ \\
\bottomrule
\end{tabular}
\end{table}

\paragraph{Verification in practice.} Table~\ref{tab:verify} measures 
invariance at the representation level. We take the $3000$ sample test set of MFInstSeg \citep{wu2024aagnet} and make three different perturbations: repartitioning, NURBS re-expresson, and rigid motions. We compare the perturbed version with the original by first comparing the canonical region graph. We match the counts and then the largest feature disagreement after optimally matching regions. We show residuals in units of $\sigma$, the channel's standard deviation over the training set. We see that the invariance properties hold in practice. Appendix~\ref{app:verify} provides an in-depth analysis of the invariance.

\section{Experiments}
\label{sec:exp}

\begin{table}[t]
\centering
\small
\caption{Segmentation under automatic perturbations. We show that our method is robust to different perturbations of the datapoints across a range of benchmark segmentation tasks, while the performance of popular baselines degrades significantly.}
\label{tab:main}
\begin{tabular}{llcccccc}
\toprule
dataset & model & clean & axis-split & diag-split & rotation & NURBS & composed \\
\midrule
\midrule
MFInstSeg & AAGNet  & 0.9851 & 0.6927 & 0.4663 & 0.4811 & 0.0604 & 0.0055 \\
MFInstSeg & UV-Net         & 0.9725 & 0.5486 & 0.3001 & 0.3392 & 0.0035 & 0.0026 \\
MFInstSeg & BRepNet        & 0.9828 & 0.7176 & 0.4626 & 0.7746 & 0.9564 & 0.4428 \\
MFInstSeg & DGCNN  & 0.6622 & 0.6622 & 0.6622 & 0.1231 & 0.6622 & 0.1231 \\
MFInstSeg & \textbf{Ours}  & {\bf 0.9862} & \textbf{0.9870} & \textbf{0.9873} & \textbf{0.9867} & \textbf{0.9868} & \textbf{0.9862} \\
\midrule
MFCAD++   & AAGNet         & 0.9851 & 0.6737 & 0.4357 & 0.6027 & 0.0275 & 0.0056 \\
MFCAD++   & UV-Net         & 0.9770 & 0.5964 & 0.3569 & 0.4115 & 0.0079 & 0.0040 \\
MFCAD++   & BRepNet        & 0.9842 & 0.7593 & 0.5209 & 0.7491 & 0.9596 & 0.4661 \\
MFCAD++   & DGCNN  & 0.6571 & 0.6571 & 0.6571 & 0.1095 & 0.6571 & 0.1095 \\
MFCAD++   & \textbf{Ours}  & \textbf{0.9870} & \textbf{0.9887} & \textbf{0.9884} & \textbf{0.9851} & \textbf{0.9856} & \textbf{0.9846} \\
\midrule
CADSynth  & AAGNet         & 0.9904 & 0.4698 & 0.4499 & 0.9463 & 0.0360 & 0.0219 \\
CADSynth  & UV-Net         & 0.9859 & 0.4257 & 0.3975 & 0.9141 & 0.0250 & 0.0089 \\
CADSynth  & BRepNet        & 0.9904 & 0.6487 & 0.6212 & 0.9766 & 0.9121 & 0.7249 \\
CADSynth  & DGCNN  & 0.6142 & 0.6142 & 0.6142 & 0.2673 & 0.6142 & 0.2673 \\
CADSynth  & \textbf{Ours}  & \textbf{0.9922} & \textbf{0.9926} & \textbf{0.9930} & \textbf{0.9906} & \textbf{0.9909} & \textbf{0.9816} \\
\bottomrule
\end{tabular}
\end{table}

\begin{figure}[t]
\centering
\begin{tikzpicture}
\begin{axis}[width=0.72\linewidth, height=4.6cm, xmode=log,
  xlabel={face-count increase (\%)}, ylabel={macro mIoU},
  xtick={3.8,15.2,30.3,76.9,191.7},
  xticklabels={3.8,15.2,30.3,axis-split,diag-split},
  ymin=0.25, ymax=1.02, legend pos=south west, legend style={font=\small},
  grid=major, tick label style={font=\small}, label style={font=\small}]
\addplot[thick, mark=*, color=red!70!black] coordinates
  {(3.8,0.9657) (7.6,0.9489) (15.2,0.9118) (30.3,0.8344) (76.9,0.6927) (191.7,0.4663)};
\addlegendentry{AAGNet}
\addplot[thick, mark=triangle*, color=violet!80!black] coordinates
  {(3.8,0.9370) (7.6,0.9080) (15.2,0.8457) (30.3,0.7342) (76.9,0.5486) (191.7,0.3001)};
\addlegendentry{UV-Net}
\addplot[thick, mark=diamond*, color=teal!80!black] coordinates
  {(3.8,0.9643) (7.6,0.9474) (15.2,0.9107) (30.3,0.8406) (76.9,0.7176) (191.7,0.4626)};
\addlegendentry{BRepNet}
\addplot[thick, dotted, color=gray, mark=none, domain=3:200] {0.6622};
\addlegendentry{DGCNN (invariant)}
\addplot[thick, mark=square*, color=blue!70!black] coordinates
  {(3.8,0.9869) (7.6,0.9869) (15.2,0.9868) (30.3,0.9869) (76.9,0.9870) (191.7,0.9873)};
\addlegendentry{Ours}
\end{axis}
\end{tikzpicture}
\caption{Re-paritition intensity effect on MFInstSeg (log scale in the $x$-axis). Every B-rep baseline degrades monotonically in the amount of face splitting while ours is flat at parity accuracy. DGCNN's point-cloud
input is partition-invariant by construction, but at a massive
accuracy cost.}
\label{fig:dose}
\end{figure}
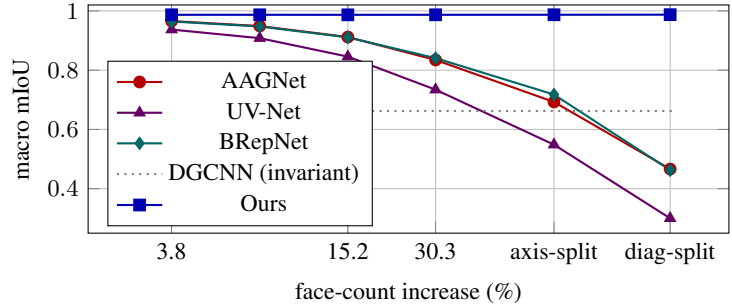

We compare our method with popular baselines UV-Net~\citep{jayaraman2021uv}, AAGNet~\citep{wu2024aagnet} and BrepNET~\citep{lambourne2021brepnet}. We also include a point cloud method DGCNN~\citep{wang2019dynamic}, which does not depend on the B-rep (but less accurate because the point clouds give less information than the B-rep). We use each baseline's published
weights where they exist and retrain at a shared budget where they do not. Full details in Appendix~\ref{app:impl}.


There are several ways to automatically generate different valid B-reps for a solid. We explore:
\begin{itemize}
    \item {\bf axis-split.} Re-partitioning each face by sectioning two axis-aligned planes 
\item {\bf diag-split.} Re-partitioning each face is sectioned by four planes including diagonals
\item {\bf split-k.} Re-partitioning where exactly $k$ faces are split.
\item {\bf rotation.} Applies one random rigid motion per part, shared by every model.
\item {\bf NURBS re-expression.} Rewrite each analytic surface as its
exact B-spline form.
\item {\bf composed.} Applies diag-split, rotation and re-expression at once.
\end{itemize}

{\bf Segmentation Benchmark}: We use standard benchmark datasets and compare the accuracy on the original task vs accuracy on a task where the datapoints are perturbed automatically.
Table~\ref{tab:main} shows the results. All accuracy numbers are macro mIoU (per-class intersection-over-union of predicted vs. true face labels), averaged over classes with $1.0$ being the perfect score. For each perturbation, the geometry of the solid remains unchanged (up to a very small tolerance). We see that our method has robust performance under these perturbations, while the popular baselines' performances decay significantly. It's worth noting that DGCNN consumes point clouds sampled from the geometry, so re-partitioning and re-expression remains unchanged but is affected by rotation. Figure~\ref{fig:dose} shows how performance monotonically decays for the B-rep baselines as we increase the amount of re-partitioning, while our method remains stable.

\begin{table}[t]
\centering
\small
\caption{Retrieval self-identification on $3000$ real Fusion~360 parts.
We report the rank-1 accuracy of retrieving a part's own clean B-rep (with retrieval database consisting of all the clean B-reps) from a query that is the same solid under one perturbation. DGCNN's queries are
re-sampled point clouds so there is some variance even in the columns it's theoretically invariant to.}
\label{tab:retrieval}
\setlength{\tabcolsep}{4.5pt}
\begin{tabular}{lccccc}
\toprule
model & axis-split & diag-split & rotation & NURBS & composed \\
\midrule
AAGNet  & 22.6\% & 6.3\% & 25.5\% & 0.7\% & 0.2\% \\
UV-Net  & 5.5\%  & 1.0\% & 7.3\%  & 0.3\% & 0.1\% \\
BRepNet & 10.6\% & 7.8\% & 16.5\% & \textbf{98.1\%} & 3.6\% \\
DGCNN   & 85.9\% & 85.6\% & 7.7\% & 86.3\% & 7.8\% \\
\textbf{Ours} & \textbf{96.1\%} & \textbf{96.3\%} & \textbf{88.3\%} & 80.5\% & \textbf{76.1\%} \\
\bottomrule
\end{tabular}
\end{table}

\begin{table}[t]
\centering
\small
\caption{Predictive churn on $3{,}000$ MFInstSeg solids. Percentage of
individual face predictions that change with perturbation, and the fraction of solids with at least one change.}
\label{tab:paired}
\setlength{\tabcolsep}{4.5pt}
\begin{tabular}{lccccc}
\toprule
 & axis-split & diag-split & rotation & NURBS & composed \\
\midrule
\multicolumn{6}{l}{\emph{faces flipped}} \\
AAGNet  & $11.90\%$ & $27.29\%$ & $35.83\%$ & $74.68\%$ & $96.81\%$ \\
UV-Net  & $21.66\%$ & $45.69\%$ & $50.29\%$ & $97.80\%$ & $98.19\%$ \\
BRepNet & $10.75\%$ & $24.45\%$ & $8.98\%$  & $1.20\%$  & $29.90\%$ \\
\textbf{Ours} & $\textbf{0.00\%}$ & $\textbf{0.00\%}$ & $\textbf{0.00\%}$ & $\textbf{0.00\%}$ & $\textbf{0.02\%}$ \\
\midrule
\multicolumn{6}{l}{\emph{solids affected}} \\
AAGNet  & $85.63\%$ & $100.00\%$ & $96.77\%$ & $99.93\%$ & $100.00\%$ \\
UV-Net  & $96.00\%$ & $100.00\%$ & $99.83\%$ & $100.00\%$ & $100.00\%$ \\
BRepNet & $86.00\%$ & $99.30\%$  & $64.10\%$ & $20.87\%$ & $99.10\%$ \\
\textbf{Ours} & $\textbf{0.00\%}$ & $\textbf{0.00\%}$ & $\textbf{0.00\%}$ & $\textbf{0.03\%}$ & $\textbf{0.33\%}$ \\
\bottomrule
\end{tabular}
\end{table}

{\bf Retrieval Benchmark}: We build a label-free retrieval benchmark from
$3000$ real Fusion~360 Gallery parts. The retrieval database consists of the clean B-reps, and each query is the \emph{same} solid under one perturbation. No retrieval
training is performed. Table~\ref{tab:retrieval} shows the results. BRepNet
is best with NURBS perturbation as re-expression changes nothing its
features read, so its embeddings barely move but performs poorly under other perturbations. Ours is the only model that consistently performs across perturbations.

\paragraph{Predictive Churn.} It is quite common for models to have very different individual predictions, even with similar accuracy. This is known as predictive churn \citep{jiang2021churn}; which can be seen as a more direct measure of stability than for e.g. accuracy on downstream tasks. We see in Table~\ref{tab:paired} that our method has very low churn while the baselines all exhibit a high amount, sometimes near $100\%$ in terms of number of face predictions changed, a clear sign of representation collapse. 

\paragraph{Why augmentation is not the answer.} A natural approach is to train with the perturbations applied via data augmentation. Table~\ref{tab:aug} shows the results with each baseline trained with each of the augmentations, evaluated on every perturbation. Augmentation genuinely helps and in all cases, substantially improves the results for its corresponding perturbation; however, the transfer is not as strong across different perturbations. Furthermore in all cases, our method without any data augmentation surpasses the performance on each individual perturbation, suggesting that our method of fixing the problem at the architecture level is superior to augmenting the data.

\begin{figure}[t]
\centering
\begin{tikzpicture}
\begin{axis}[width=0.86\linewidth, height=3.5cm, ybar, bar width=16pt,
  symbolic x coords={AAGNet, UV-Net, BRepNet, Ours},
  xtick=data, ymin=0, ymax=130, ytick={0,25,50,75,100},
  ylabel={rank-1 retrieval (\%)}, ylabel near ticks,
  nodes near coords, nodes near coords style={font=\footnotesize,
    /pgf/number format/.cd, fixed, precision=1, zerofill},
  every axis plot/.append style={fill=gray!45, draw=gray!70!black},
  tick label style={font=\small}, enlarge x limits=0.14]
\addplot coordinates {(AAGNet,57.0) (UV-Net,18.3) (BRepNet,23.7)
  (Ours,98.1)};
\end{axis}
\end{tikzpicture}
\caption{Kernel-induced re-partitioning in OpenCASCADE: rank-1 retrieval accuracy.}
\label{tab:genvar}
\end{figure}
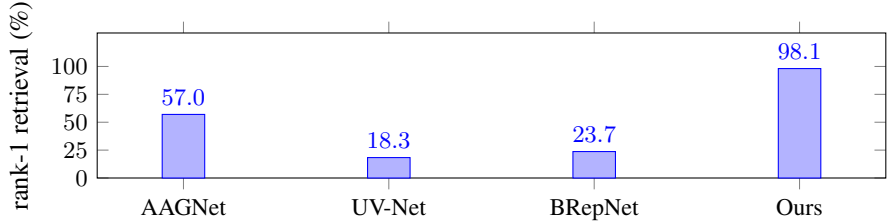

\paragraph{Kernel-induced re-partitioning.} In our automatic perturbations based on re-partitioning, we use \texttt{SplitShape} command in OpenCASCADE on the B-rep. We show that our results also hold when we don't use a splitting operator and instead rewrite the CAD program itself in OpenCASCADE in the following way: whenever a sketch is extruded by extent E, we make the same sketch extruded E/2, plus a second copy on a plane offset by E/2, extruded E/2 so in the end it is still the same solid but now each extrusion is partitioned into two. The dataset we use is $1{,}049$ pairs, generated procedurally based on CAD-Recode \citep{rukhovich2025cad}. Figure~\ref{tab:genvar} shows the results and our method outperforms significantly.

\paragraph{Region Ablation} In Table~\ref{tab:ablation_region}, we show the importance of face merging in our region-based approach. We compare the robustness under our automatic re-partitioning perturbation using an input graph with the region-based nodes (with merging) vs the input graph without merging (using the original faces as nodes). 

\begin{table}[t]
\centering
\small
\caption{Data augmentation matrix: every baseline and every
augmentation, evaluated on every perturbation. We report macro mIoU on MFInstSeg test set.}
\label{tab:aug}
\setlength{\tabcolsep}{4.5pt}
\begin{tabular}{lccccc}
\toprule
model & clean & axis-split & diag-split & NURBS & composed \\
\midrule
AAGNet $+$ axis aug.\        & 0.9850 & 0.9759 & 0.8558 & 0.0450 & 0.0054 \\
\ $+$ diag aug.\        & 0.9828 & 0.9774 & 0.9703 & 0.0469 & 0.0043 \\
\ $+$ NURBS aug.\       & 0.9851 & 0.6778 & 0.4711 & 0.9853 & 0.2358 \\
\ $+$ composed aug.\    & 0.9802 & 0.9421 & 0.5725 & 0.9674 & 0.9502 \\
\midrule
UV-Net $+$ axis aug.\   & 0.9876 & 0.9630 & 0.7622 & 0.0072 & 0.0066 \\
\ $+$ diag aug.\        & 0.9827 & 0.9561 & 0.9421 & 0.0104 & 0.0053 \\
\ $+$ NURBS aug.\       & 0.9744 & 0.5420 & 0.3108 & 0.9744 & 0.1893 \\
\ $+$ composed aug.\    & 0.9717 & 0.8152 & 0.5083 & 0.9456 & 0.9326 \\
\midrule
BRepNet $+$ axis aug.\  & 0.9849 & 0.9830 & 0.9192 & 0.9587 & 0.3037 \\
\ $+$ diag aug.\        & 0.9828 & 0.9793 & 0.9780 & 0.9565 & 0.6115 \\
\ $+$ NURBS aug.\       & 0.9785 & 0.7238 & 0.4752 & 0.9588 & 0.5832 \\
\ $+$ composed aug.\    & 0.9816 & 0.9740 & 0.8125 & 0.9586 & 0.9639 \\
\midrule
\textbf{Ours (no aug.)} & 0.9862 & \textbf{0.9870} & \textbf{0.9873} & \textbf{0.9868} & \textbf{0.9862} \\ \bottomrule
\end{tabular}
\end{table}

\begin{table}[t]
\centering
\caption{Region ablation: we keep features the same but test merging (into regions) vs not merging (keeping the existing faces) for the input graph.
 We use MFInstSeg test split under
axis/diag re-partitioning.
``Solids stable'' is the fraction of test solids whose predictions are
bit-identical between the clean file and its re-partitioned version. 
}
\label{tab:ablation_region}
\setlength{\tabcolsep}{4pt}
\begin{tabular}{lcccc}
\toprule
& \multicolumn{2}{c}{mIoU} & \multicolumn{2}{c}{solids stable} \\
& axis & diag & axis & diag \\
\midrule
face nodes (merging off)   & 0.5653 & 0.3502 & 4.80\% & 0.07\% \\
region nodes (merging on)  & \textbf{0.9828} & \textbf{0.9840} & \textbf{99.73\%} & \textbf{99.70\%} \\
\bottomrule
\end{tabular}
\end{table}

\begin{table}[t]
\centering
\caption{Robustness to an ordinary Rhino import/export of the entire MFInstSeg test split. We show macro mIoU as in Table~\ref{tab:main} and additionally the predictive churn at both face-level and solid-level (if at least one face's prediction in the solid was changed).}
\label{tab:rhino}
\small\setlength{\tabcolsep}{5pt}
\begin{tabular}{lcccc}
\toprule
 & clean & round-trip & faces flipped & solids affected \\
\midrule
AAGNet (publ.) & 0.9851 & 0.2838 & $47.04\%$ & $99.50\%$ \\
UV-Net         & 0.9725 & 0.0433 & $89.98\%$ & $100.00\%$ \\
BRepNet        & 0.9828 & 0.9674 & $0.37\%$ & $5.13\%$ \\
\textbf{Ours}  & 0.9862 & \textbf{0.9870} & $\textbf{0.00\%}$ & $\textbf{0.04\%}$ \\
\bottomrule
\end{tabular}
\end{table}

\begin{table}[t]
\centering
\caption{The human study: segmentation predictive churn (percentage) between two independently authored FreeCAD scripts of the same geometry.}
\label{tab:human}\small
\begin{tabular}{lcccc}
\toprule
 & AAGNet & UV-Net & BRepNet & Ours \\
\midrule
scripts from same expert & 11.8 & 30.1 & 12.6 & \textbf{0.5} \\
scripts from different experts   & 13.1 & 15.1 & 12.5 & \textbf{0.2} \\
\bottomrule
\end{tabular}
\end{table}

\begin{table}[t]
\centering
\caption{Fusion~360 under the automatic perturbations, same conventions
as Table~\ref{tab:main}}
\label{tab:f360}\small
\setlength{\tabcolsep}{4.5pt}
\begin{tabular}{lcccccc}
\toprule
model & clean & axis-split & diag-split & rotation & NURBS & composed \\
\midrule
AAGNet         & \textbf{0.7412} & 0.4842 & 0.3151 & 0.5512 & 0.2473 & 0.0697 \\
UV-Net         & 0.6977 & 0.3772 & 0.2218 & 0.4576 & 0.2895 & 0.0950 \\
BRepNet        & 0.7162 & 0.3541 & 0.2419 & 0.5167 & 0.5707 & 0.2415 \\
DGCNN (points) & 0.4025 & 0.4025 & 0.4025 & 0.1436 & 0.4025 & 0.1436 \\
\textbf{Ours}  & 0.6148 & \textbf{0.6048} & \textbf{0.5725} & \textbf{0.6153} & \textbf{0.6150} & \textbf{0.6142} \\
\bottomrule
\end{tabular}
\end{table}

\paragraph{Kernel round-trips.} Here we explore the effect of a real-world perturbation: a plain import/export round-trip in Rhino, a popular commercial CAD software which uses the openNURBS kernel. Table~\ref{tab:rhino} shows the results: AAGNet and UV-Net collapse catastrophically, while BRepNet holds better: its features do not read surface parameterizations. Our method is still the most stable.

\paragraph{Human study.} The human study runs in FreeCAD where we have two experts who labeled $25$ parts. One labeler made $3$ different FreeCAD scripts for each part and a second labeler made one script for each part. We compare both variation in scripts made by the same expert and variation made by different experts. The results are in Table~\ref{tab:human}, and full details are in Appendix~\ref{app:human}. We compare the methods on the segmentation predictive churn on each methods' corresponding MFInstSeg-trained checkpoint for both within-expert and cross-expert B-rep pairs. We see that our method is far more consistent than the baselines.

\paragraph{The cost of invariance (a limitation).} Our method has a limitation in situations the information in the B-rep itself can be useful and not using it would hurt performance. One such situation is the Fusion~360 Gallery segmentation, where it labels each face by the modeling
operation that created it. As such, a partition-invariant model cannot always recover such labels, while a model free to read the boundary
decomposition can exploit residual traces of the history. Table~\ref{tab:f360} shows the results. We do see that our method performs worse than the B-rep baselines on the clean dataset; however, when automatic perturbations are applied, our robustness overcomes the decay in the baselines when faced with the perturbations.

\section{Conclusion}

We showed that existing B-rep encoders are not robust to variation in the B-rep used to represent the solid. We proposed the \emph{canonical region graph}, an input representation whose nodes, features and coordinate frame are derived from the solid itself with theoretical invariance guarantees.  It matches the strongest baseline on standard benchmarks and is stable under variation of B-rep.

Our findings may have implications for B-rep generators that count re-descriptions of the same solid as distinct or novel samples \citep{xu2024brepgen,lee2025brepdiff,xu2025autobrep,li2025dtgbrepgen,li2025brepgpt} and those that use element-matching reconstruction scores \citep{guo2022complexgen}, which can penalize correct solids decomposed differently from the one reference B-rep.

Future work includes extending the method to use feature trees and construction histories but with invariance constraints in order to leverage potentially useful information contained in the B-rep. Another direction is extending the method to work for assemblies, which are files with multiple parts, instead of just a single part, as well as 2D sketches and drawings.

\bibliography{references}
\bibliographystyle{paper}

\newpage
\appendix

\section{Proofs}
\label{app:proofs}

\paragraph{Proof of Proposition 1.} Analyticity gives uniqueness of the surface
decomposition: two distinct analytic surfaces agree at most on a set with
empty interior in either, so every point of $\partial S$ off a measure-zero
seam set lies in exactly one maximal surface, and the patches are the
connected components $\Sigma_1,\dots,\Sigma_m$ of the corresponding subsets
of $\partial S$. Let $P$ be a valid partition. Each face lies in exactly one
$\Sigma_j$, and the faces contained in $\Sigma_j$ cover it with disjoint
interiors. It remains to show the edge-adjacency graph of the faces covering
one connected $\Sigma_j$ is connected: for $x,y$ in the interiors of two such
faces, take a path in $\Sigma_j$ avoiding the (finitely many) face vertices;
the path crosses face boundaries finitely often, each crossing at an interior
point of a shared piecewise-regular curve, i.e.\ a positive-length shared
edge. The sequence of faces along the path is a walk in the adjacency graph.
Hence regions $=$ patches, for every $P$; region adjacencies and shared
boundary curves are likewise determined by the patches alone. \hfill$\square$

\paragraph{Proof of Proposition 2.} Throughout, $\varphi$ denotes an arbitrary
integrable function on $\partial S$ (the integrand of any surface feature) and
$dA$ the surface area element. If region $R$ is cut into faces $f_1,\dots,f_k$
with disjoint interiors, then $\int_R \varphi\,dA=\sum_i \int_{f_i}\varphi\,dA$
because seams have measure zero. For boundary features --- integrals
$\int_\Gamma \psi\,d\ell$ of a function $\psi$ along a boundary curve
$\Gamma$, $d\ell$ the arc-length element: by Proposition~1,
$\partial R$ and each shared boundary $\partial R \cap \partial R'$ ($R'$
ranging over the other regions) are
partition-independent sets; internal seams introduced by splitting separate
faces of the \emph{same} region and are excluded by construction, so
boundary integrals also decompose over faces. \hfill$\square$

\paragraph{Proof of Proposition 3.} Partition-invariance of a predictor $f$
means $f(x)=f(x')$ whenever its inputs
$x,x'$ are B-reps of the same solid, i.e.\ $f$ factors through the geometry
map $g$ --- the map sending a B-rep to the solid it bounds (unrelated to the
rigid motion $g$ of Proposition~5). For any such $f$, a B-rep-valued random
input $X$, and its label $Y$,
$\Pr[f(X)\ne Y] = \mathbb{E}\big[\Pr[f \ne Y \mid g(X)]\big]
\ge \mathbb{E}\big[\min_c \Pr[Y \ne c \mid g(X)]\big]$,
the Bayes risk of predicting $Y$ from geometry alone (the minimum ranging
over label values $c$). If two histories with
different labels produce the same solid with positive probability, this bound
is strictly positive; conversely any predictor with lower risk cannot factor
through $g$ and therefore depends on the representation of $\partial S$ ---
the partition. \hfill$\square$

\paragraph{Proof of Proposition 4.} Write $C(S)$ for the boundary's area
centroid, $I_C(S)$ for its inertia tensor, $V(S)$ for the matrix of
principal axes (eigenvectors of $I_C$), and $g=(\mathbf{R},\mathbf{t})$
for a rigid motion --- rotation $\mathbf{R}$, translation $\mathbf{t}$
--- with $gS$ the moved solid. Translation: the centroid is
area-mean of a boundary integral, so $C(gS)=\mathbf{R}C(S)+\mathbf{t}$ is
immediate; centred coordinates $x-C$ eliminate $\mathbf{t}$. Rotation: for
the centred inertia $I_C = \int \big(\|x-C\|^2 E - (x-C)(x-C)^{\!\top}\big) dA$
--- $x$ ranging over $\partial S$ and $E$ the $3\times3$ identity matrix ---
substituting $x \mapsto \mathbf{R}x+\mathbf{t}$ gives
$I_C(gS)=\mathbf{R}I_C(S)\mathbf{R}^{\!\top}$, whose eigenvectors are
$\mathbf{R}V$ with the same eigenvalues. The residual ambiguity group is
$\{\pm 1\}^3$ on signs times $SO(2)$ (rotations within a plane; $SO(3)$,
all 3D rotations, for triple degeneracy) within each degenerate
eigenspace; the sign choice is a function of the third moments --- the integrals
$\int \langle v_j,\,x-C\rangle^3\,dA$ along the principal axes $v_j$,
invariants after centring and rotation --- and the degenerate-subspace average is by
construction constant on orbits of the ambiguity group. Features that are
tensors expressed in this frame are therefore fixed points of the whole
nuisance group. Partition-invariance is inherited because every ingredient
--- $C$, $I_C$, the third moments --- is an integral over $\partial S$
(Proposition~2). An improper orthogonal transformation can reverse the handedness of the skew-oriented eigenframe. Independent sign fixing does not by itself preserve uniqueness under reflection, because coordinates expressed in the corresponding reflected frame can coincide. The frame is therefore constrained to be right-handed: when the sign fixes land left-handed, the determinant correction is distributed over the near-tied smallest-skew axes by continuously reweighting their sign blends (reducing to a single axis flip when the minimum is unique). Mirror pairs are then distinguished whenever all three skew statistics exceed the blend band, the smallest skew magnitude is untied, and the inertia spectrum is simple; at a tie or inside the band, discrimination degrades continuously to zero and is not guaranteed.
\hfill$\square$

\paragraph{Proof of Lemma 1.} Soft binning assigns mass $m$ at coordinate $t$ to bins
$\lfloor t\rfloor,\lfloor t\rfloor{+}1$ with weights $1{-}\{t\},\{t\}$,
writing $\lfloor t\rfloor$ for the integer part and $\{t\}$ for the
fractional part of $t$; here $w$ is the bin width, coordinates are measured
in units of $w$, and in our histograms the mass $m$ is the sample's
quadrature weight --- its triangle's area (its segment's length for the
boundary histograms). The map $t \mapsto$ (weight vector) is piecewise linear
with slope $\pm 1/w$ into two coordinates, so a perturbation $\delta$ of one
sample moves the histogram by at most $2m|\delta|/w$ in $L_1$; summing over
samples gives the bound. The histogram is also linear in the masses, each
mass spreading with total weight one, so perturbing an area weight by
$\delta$ moves it by at most $|\delta|$ in $L_1$: Lipschitz constant $1$ in
the weights. The remaining mesh-computed features --- the integrated normal
and the curvature totals --- are plain quadrature sums
$\sum_T a_T\,\varphi(x_T)$ over triangles $T$ with area weights $a_T$,
sample positions $x_T$, and integrand $\varphi$ (unit normals; curvatures):
linear in the $a_T$ with coefficients bounded by $\sup|\varphi|$, and
Lipschitz in the $x_T$ with constant $\mathrm{Lip}(\varphi)\sum_T a_T$,
where $\mathrm{Lip}(\varphi)$ --- finite because each integrand is smooth on
the fixed analytic surface the samples lie on --- is the integrand's
Lipschitz constant there. Grid blending is Lipschitz in the blend weight for fixed binnings; this does not establish continuity across spatial bin boundaries. The histogram and quadrature bounds above establish the result for these features. Neither hard variant admits any
bound: hard binning moves a sample's full mass $m$ between bins when a
perturbation $\delta\to0$ crosses a bin edge (an $L_1$ jump of $2m$
regardless of $\delta$), and a hard sign choice ($w \in \{0,1\}$)
likewise has no finite Lipschitz constant --- this is why the rotation
row of Table~\ref{tab:verify} concentrates its residual in histogram
channels. \hfill$\square$

\section{Experiment implementation details}
\label{app:impl}

\paragraph{Perturbation construction.} Re-partitioning uses
\texttt{BRepFeat\_SplitShape} along section curves that lie on the faces they
cut. A split face's children inherit its label. The acceptance tolerance quoted throughout as ``a very small tolerance'' is relative and unit-free: a
perturbed solid is kept only if $|\Delta V|/V \le 10^{-9}$ and
$|\Delta A|/A \le 10^{-9}$, i.e.\ volume (V) and surface area (A) each match the original to within one part in $10^{9}$, independent of the model's units. Every other tolerance in the paper is relative in the same sense unless stated otherwise.

\paragraph{Rotation and re-expression.} Rotation draws a uniform random
axis and an angle uniform in $[0.2, \pi]$ and every model sees the identical
rotated solid. Re-expression uses
OpenCASCADE's \texttt{NurbsConvert}. Labels transfer to perturbed files by index where face order survives (rotation) and by multi-point interior on-surface matching with an unambiguity requirement; faces with no unambiguous parent are explicitly rejected and excluded from the metric identically for every model, with rejected-face counts and surface-area share reported per dataset and perturbation (with the clean solid first moved by the same rotation, so both solids sit in the
same model-space coordinates). A small
fraction of perturbed files crash some baseline's own feature extractor;
such parts are dropped from that column for all models.

\paragraph{Training.}  For our models: we use AdamW, lr $10^{-3}$,
a OneCycleLR schedule, batch 64, label smoothing $0.05$, EMA $0.999$; $d{=}512$,
$8$ EdgeGAT layers, $8$ heads ($17.5$M parameters). Synthetic benchmarks
train $80$ epochs; Fusion~360 trains $200$ epochs with inverse-frequency
class weights at exponent $0.3$. Released checkpoints retain their original training protocols; retrained baselines are identified separately.

\paragraph{Evaluation.} We evaluate on the official test split, and only for examples whose perturbations pass the validity test. Checkpoints were always selected by validation, but the choice among the Fusion~360 feature variants involved repeated test evaluations and validation-based selection would pick a variant within $0.3$ mIoU of the top performing result. DGCNN on MFInstSeg and
Fusion~360 uses a configuration of $4096$ points and doubled
epochs. We train and evaluate our method on three random initializations, trained once on that dataset's official training split, and for the  baselines we use their 
released weights (or one retraining at the shared budget where none are
released). Every number in the paper, on every perturbation and study, is
an inference-mode evaluation of these fixed checkpoints. Epoch checkpoints were selected by validation mIoU. Feature-variant development on Fusion 360 involved repeated test evaluations, so its test results should not be interpreted as an untouched development holdout. After fixing the representation and training protocol, we trained three random initializations per dataset. The main tables report the run selected by validation mIoU. Baselines use released weights where available and otherwise a retraining under the stated protocol. Reported evaluations use fixed checkpoints in inference mode, with no further adaptation during evaluation. DGCNN on MFInstSeg and Fusion 360 uses 4,096 points and twice the stated epoch budget.

\paragraph{Retrieval embeddings.} We use mean$+$max pools of each model's own
node features, taken from the released checkpoints without any retrieval
training; we cosine similarity on the embeddings to find the top-1 in the database.

\section{The human study in detail}
\label{app:human}
\begin{table}[t]
\centering
\caption{The human study's full results for predictive churn: the percentage of faces whose predicted label changes between
two independently authored constructions of the same geometry. Upper block:
within-designer pairs. Lower block: the between-designer. }
\label{tab:human_full}
\small\setlength{\tabcolsep}{4pt}
\begin{tabular}{lcccc}
\toprule
 & AAGNet & UV-Net & BRepNet & Ours \\
\midrule
geometry-exact pairs ($87$ of them) & 11.8 & 30.1 & 12.6 & \textbf{0.5} \\
all $113$ pairs (bodies pooled)      & 11.7 & 27.5 & 11.1 & \textbf{0.6} \\
raw file vs.\ its canonical twin     & 11.4 & 12.1 & 9.0 & \textbf{0.2} \\
raw cross-version pairs              & 13.5 & 35.9 & 13.1 & \textbf{0.3} \\
\midrule
cross-designer geometry-exact pairs ($24$ of them) & 13.1 & 15.1 & 12.5 & \textbf{0.2} \\
2nd designer's solids, axis-split    & 18.9 & 64.1 & 30.6 & \textbf{0.2} \\
2nd designer's solids, diag-split    & 27.4 & 75.4 & 39.8 & \textbf{0.0} \\
unchanged faces across designers     & 15.7 & 18.9 & 15.0 & \textbf{0.7} \\
\bottomrule
\end{tabular}
\end{table}

\paragraph{Calibration.} Before the study we checked how a designer's habits reach the exported file. They do indeed affect the exported file: e.g. draw a circular
profile as two half-arcs and the exported cylindrical face arrives split
in two; model half a part and mirror it, and the mirror plane leaves extra
seam faces. One pass of the standard cleanup (\texttt{refine}, OCC's
\texttt{UnifySameDomain}) removed many artifacts, and the kernel normalizes some route choices before they even
reach the file (building a cylinder by lofting between two circles stores
the same analytic cylinder as extruding one). Because much of the software
stack applies such cleanup automatically, habit-level variation rarely
reaches a consumer; the study therefore cleans every file first, and the
deployment-scale measurements target the paths where no cleanup
intervenes i.e. kernels that store everything as NURBS (Rhino's openNURBS)
and translation between kernels.

\paragraph{Multi-body intake.} Some parts are built as
several separate pieces rather than one solid, and two files of the same
part need not list those pieces in the same order. Pieces are
therefore matched by size and position (volume and centroid). Evaluating
piece by piece and pooling the per-face results recovers every program
the one-solid intake rejects ($27$ of the $100$ are genuinely
multi-body), so all $100$ human programs are usable.

\paragraph{Staging and audit.} All files from both experts are
produced by executing their $100$ FreeCAD programs through one deterministic
runner. Files come from people re-typing dimensions off drawings, and a single mistyped value makes the
two solids genuinely different shape. Sameness is therefore tested per pair. Pairs are gated by a spatial identity check: maximum bidirectional deviation of interior on-surface sample points within 1e-6 relative (agreement at the stated sampling resolution, not an exact certificate) — with volume/area agreement at 1e-6 as a confirmatory scalar check. Scalars alone are insufficient: a feature moved inside a part can leave every scalar unchanged. 
Among $122$ pairs satisfying the scalar gate, checks at three sampling densities confirm $116$ within the stated spatial tolerance; three are tolerance-boundary cases and three are geometrically different. The confirmed set contains $92$ within-designer pairs and $24$ cross-designer pairs. The primary comparison further requires successful predictions from every compared model, leaving $87$ within-designer pairs and all $24$ cross-designer pairs. Boundary cases, different geometries, and model-specific extraction failures are reported separately.

\paragraph{Scoring support.}
Consistency for every model is computed on the same face support: faces
whose cross-construction correspondence is accepted (ambiguous or
off-surface transfers are excluded identically for all models),
restricted to pairs on which every compared model produces a prediction.
Five within-expert pairs (two designs) are unparseable by the baselines'
own feature extractors; they are excluded from the paired comparison but
reported, and per-model coverage accompanies each table. 

\paragraph{Between-designer geometric identity.} Free modeling from a
drawing does not produce geometrically identical solids. The experts were instructed to match the output dimensions after making a first pass independently. Of the $35$ staged cross-designer pairs, there were $26$ pairs whose volume and surface area agree within the specified tolerance. Then checking additional points on both models’ surfaces, we classified $24$ as confirmed within tolerance, one as tolerance-boundary, and one as geometrically different.

\paragraph{Derived measurements.} Re-partitioning the second user's own
solids exercises human-authored geometry under our synthetic families. The results are in Table~\ref{tab:human_full}.

\paragraph{The unchanged-face cell.} For pairs whose
geometry does not match exactly, consistency is scored only on the faces
both users built identically (matched at $10^{-6}$ in surface type, area
and centroid). One caveat applies to every model equally: all of these
are graph networks, so a prediction on an unchanged face may legitimately
move because the \emph{rest} of the part differs. This is
therefore a softer test.

\section{Canonical frame details}
\label{app:rigid}

Some of the model's features describe placement i.e. where a
region sits and which way it is oriented. Coordinates require a reference
frame, and the file's own frame is arbitrary: rotate the part and every
such feature changes, which is exactly the rotation sensitivity the
baselines exhibit. We compute the canonical frame from
the solid itself: the eigenvectors of its boundary inertia tensor,
which is built from integrals and is therefore partition-invariant too.

An inertia frame, however, is not automatically well-defined. Eigenvectors
come with two ambiguities: each axis is defined only up to sign, and when
two eigenvalues are (nearly) equal the axes are defined only up to a
rotation within that subspace. We resolve 
both. Axis signs are fixed by the solid's third-order moments (its skew
along each axis), exactly with a closed-form cubic over
the mesh triangles. When a
skew is too close to the threshold, the two sign choices are blended
continuously instead of hard-switched; when eigenvalues are nearly equal, frame-dependent features are blended
toward their exact averages over rotations of the degenerate subspace:
with per-gap degeneracy weights $a$ and $b$, a feature $X$ is replaced by
the convex combination
\[
(1-a)\cdot (1-b)\cdot \,X \;+\; a\cdot (1-b)\cdot \,P_{01}(X) \;+\; (1-a)\cdot b\cdot \,P_{12}(X) \;+\; a\cdot b\cdot \,P_{3}(X),
\]
where $P_{01}$ and $P_{12}$ are the closed-form averages over rotations
within each eigenvalue pair's plane and $P_{3}$ is the full rotational
average ($\operatorname{tr}(X)/3\cdot I$ for tensors, $0$ for vectors).
The projections are applied independently to the original value (composing them sequentially is order-dependent when both gaps are small) so the result is exactly basis-independent at any full degeneracy,
pairwise or triple, and continuous in the eigenvalue gaps. A genuinely symmetric solid has no preferred frame and correctly
receives none.

One further ambiguity remains: the sign fixes alone can land on either a right- or left-handed set of axes, and a mirrored solid would then receive the mirrored frame — and identical features. We therefore constrain the frame to be right-handed. When the sign fixes land left-handed, the correction is applied on the axis whose skew is least trusted, and when several axes are nearly tied for that role the correction is blended across them continuously, the same way the sign choices themselves are blended. Mirror pairs are consequently distinguished when all three skews are confident, the least-trusted one is untied, and the inertia spectrum is non-degenerate; otherwise discrimination degrades continuously and is not guaranteed (near-ties occur on 4.3\% of MFInstSeg parts).

\section{The procedure in full}
\label{app:construction}

\paragraph{Type selection on ambiguous patches.}\label{app:edge} A patch's samples do not always determine its
surface type uniquely: more than one analytic fit can clear the
acceptance tolerance. The simplest admissible type wins i.e. the one with the fewest defining
parameters (plane, then sphere, cylinder, cone, torus). Fits meeting the machine-precision threshold take priority; multiple qualifying fits are resolved by the deterministic type-selection rule.

\begin{center}
\footnotesize\setlength{\tabcolsep}{5pt}
\begin{tabular}{lp{10.4cm}}
\toprule
stage & what it does \\
\midrule
rescale & the solid is uniformly scaled to unit boundary area (factor
$1/\sqrt{A}$), so all later tolerances are size- and unit-independent;
uniform scaling is rotation-invariant, unlike the bounding-box
normalization the baselines use \\
surface keys & every face is labelled by the surface it lies on. For the
five analytic types the label is the handful of numbers that define the
surface (a cylinder's axis and radius, a plane's normal and offset),
written in a unique standard form and rounded to $10^{-6}$ so the same
surface always produces the same label. Kernels often re-save a simple
surface in the generic freeform (NURBS) format --- the way $1/2$ can be
written $0.5000$ --- so every freeform face is first tested against the five
types by fitting sampled points and normals: a fit is accepted below
$10^{-6}\!\cdot\!\mathrm{scale}$ residual (genuine matches measure
$\sim\!10^{-9}$, non-matches $O(1)$), and the face is then labelled as
its analytic self; truly freeform surfaces are labelled by their defining
control points \\
regions & faces are grouped transitively into \emph{regions}: two faces
join the same region when they share an edge and carry the same surface
label. Regions are the graph's nodes; a face's prediction is its
region's prediction \\
canonical frame & positions and directions are expressed in the solid's
own natural axes (the principal axes of its boundary inertia, as in
mechanics) rather than the file's arbitrary frame. Each axis's sign is
fixed by the solid's asymmetry (third-order moments); when an axis's
asymmetry is too small to trust (skew within $0.01$--$0.04$) the two
choices are blended, and when two axes are interchangeable
(near-equal inertia) features are blended toward the exact rotational average over the
ambiguous subspace; mirror
images are deliberately kept distinct \\
node features ($115$) & per region: which surface type it is and its
defining numbers; exact integrals over the region (area, centroid,
average normal direction, spread of positions and normals, total
curvatures); how many boundary loops it has and how long they are; how
much of its border is sharp, concave, convex; a smoothed $8$-bin
histogram of the bend angle along its border; and a coarse
$4^3$ map of where its area sits within its own bounding box \\
edge features ($10$) & per pair of touching regions: how long the shared
border is, the average and distribution ($6$ soft bins) of the bend angle
across it, and how much of it is sharp or concave \\
quadrature & all sampled quantities are computed on the kernel's triangle
mesh, refined to at most $2$M triangles per solid; histogram bins overlap
smoothly so a value near a bin edge changes features continuously
(Lemma~1) \\
network & a standard graph network over the region graph: $8$ layers of
edge-conditioned attention, width $512$, $8$ heads, dropout $0.1$; a
linear head scores each region and every face inherits its region's
score \\
\bottomrule
\end{tabular}
\end{center}

\section{Representation-level verification in full}
\label{app:verify}

This section verifies the \emph{representation} with no model involved.
For each solid and its perturbed twin, both canonical region graphs are
built and compared directly. We test whether a single bijection between the two graphs' nodes
simultaneously preserves adjacency and matches features: node counts
first, then a feature-matched assignment checked as a permuted edge set
with per-edge feature residuals. Where symmetric, feature-identical
nodes make that assignment fail adjacency spuriously, a colored
graph-isomorphism search decides whether \emph{any} structure-preserving
bijection exists, and every reported residual (node, edge, and
constant-channel drift) is then recomputed under that one common
bijection. ``Identical structure'' below means this full test passed,
not a node-count match. If the construction is invariant as claimed, structures match
and the feature differences are numerical noise. Differences are reported
in units of $\sigma$ (the feature's standard deviation across the
training set). This is the empirical backing for the propositions, measured at benchmark scale and independent of any training.

\paragraph{Feature classes.} Features split into two classes by how their
invariance is established. \emph{Exact} features (areas, moments, types,
parameters) are accumulated by the kernel's exact integrator, so their
invariance is provable and their measured drift is machine noise.
\emph{Mesh-tolerance} features (anything computed on the triangle mesh:
normal and curvature integrals, histograms, the grid) carry \emph{no
invariance guarantee}: they drift whenever a perturbation makes the kernel
re-mesh the solid or the canonical frame is recomputed. For them, the claim
is weaker and three-part: Lemma~1 bounds how fast they \emph{can}
drift, the table below measures how far they actually do, and the
experiment sections price what that drift costs at prediction level.
Only the exact class is invariant by proof. The measured drifts
below are raw relative errors on the split stress set (not $\sigma$
units). 

\begin{center}
\small\setlength{\tabcolsep}{5pt}
\begin{tabular}{llc}
\toprule
family & class & measured residual \\
\midrule
area, centroid, moments, type, axes/radii, & exact (GProp) & $\le 10^{-15}$ \\
\quad loops, boundary length & & \\
normal integrals & mesh-tolerance & $\le 1.8\times10^{-7}$ \\
curvature integrals & mesh-tolerance &  $0$  \\
dihedral histograms (node, edge) & mesh-tolerance & $\le 3\times10^{-8}$ \\
inertia-frame grid & mesh-tolerance & $\le 2.3\times10^{-5}$ \\
\bottomrule
\end{tabular}
\end{center}

\begin{table}[h]
\centering
\caption{Feature drift between each solid and its perturbed twin, by
perturbation: $N$ pairs; ``ident.''\ = the fraction whose region graphs
have identical structure; then percentiles over pairs of each
pair's \emph{largest} channel residual, in $\sigma$ units, separately for
the exact and mesh-sampled feature classes
sampled features drift where a perturbation forces a re-mesh and,
for frame-expressed channels, where the canonical frame is itself
recomputed.}
\label{tab:verify_full}
\footnotesize\setlength{\tabcolsep}{3.5pt}
\begin{tabular}{lrc|cccc|cccc}
\toprule
& & & \multicolumn{4}{c|}{exact features ($\sigma$)} & \multicolumn{4}{c}{sampled features ($\sigma$)} \\
perturbation & $N$ & ident. & p50 & p90 & p99 & worst & p50 & p90 & p99 & worst \\
\midrule
constr.\ fixtures & 24 & $24/24$ & $0$ & $0$ & $0$ & $0$ & $2$e-$15$ & $0.003$ & $0.003$ & $0.003$ \\
curved-split & 3 & $3/3$ & $0$ & $0$ & $0$ & $0$ & $0.001$ & $0.003$ & $0.003$ & $0.003$ \\
diag-split & 3000 & $99.9\%$ & $0$ & $0$ & $4$e-$7$ & $0.002$ & $0.015$ & $0.040$ & $0.40$ & $13.5$ \\
NURBS re-expr. & 3000 & $100\%$ & $0.0007$ & $0.003$ & $0.009$ & $0.022$ & $0.041$ & $0.29$ & $3.2$ & $94$ \\
random rigid motion & 3000 & $100\%$ & $0$ & $0$ & $0.002$ & $0.007$ & $4.8$ & $9.7$ & $15$ & $38$ \\
all combined & 3000 & $99.3\%$ & $0.0006$ & $0.003$ & $0.007$ & $3.2$ & $4.8$ & $9.7$ & $15$ & $80$ \\
human cross-ver. & 42 & $38/42$ & $0$ & $10^{-6}$ & $<\!10^{-4}$ & $<\!10^{-4}$ & $0.010$ & $1.9$ & $3.6$ & $3.6$ \\
\bottomrule
\end{tabular}
\end{table}

\end{document}